\documentclass[10pt,twocolumn,letterpaper]{article}

\usepackage{cvpr}
\usepackage{times}
\usepackage{epsfig}
\usepackage{graphicx}
\usepackage{amsmath}
\usepackage{amssymb}
\usepackage{authblk}
\usepackage{multirow}

\usepackage[breaklinks=true,bookmarks=false]{hyperref}

\cvprfinalcopy 

\begin{document}

\title{TextContourNet: a Flexible and Effective Framework for Improving Scene Text Detection Architecture with a Multi-task Cascade}

\author[1]{Dafang He}
\author[2]{Xiao Yang}
\author[2]{Daniel Kifer}
\author[1]{C.Lee Giles}

\affil[ ]{\textit {\{duh188,xuy111\}@psu.edu},  \textit{\{giles\}@ist.psu.edu},  \textit{dkifer@cse.psu.edu}}
\affil[1]{Information Science and Technology, The Penn State University}
\affil[2]{Computer Science and Technology, The Penn State University}

\maketitle
\begin{abstract}
  We study the problem of extracting text instance contour information from images and use it to assist scene text detection.
We propose a novel and effective framework for this and experimentally demonstrate that:
(1) A CNN that can be effectively used to extract instance-level text contour from natural images.
(2) The extracted contour information can be used for better scene text detection.
We propose two ways for learning the contour task together with the scene text detection:
(1) as an auxiliary task and (2) as multi-task cascade.
Extensive experiments with different benchmark datasets demonstrate that both designs improve the performance of a state-of-the-art scene text detector and that a multi-task cascade design achieves the best performance.
\end{abstract}

\section{Introduction}
Scene Text Detection extracts text information from natural images and has received an increasing amount of attention from both academia and industry.
Usually as an end-to-end scene text reader first detects the location of each word in the image and then a trained recognizer reads the text for each word.
For such a system, scene text detection is usually the bottleneck and much research has been devoted to improving its performance.

\begin{figure*}[!htb]
    \centering
 	\includegraphics[height=1.06in]{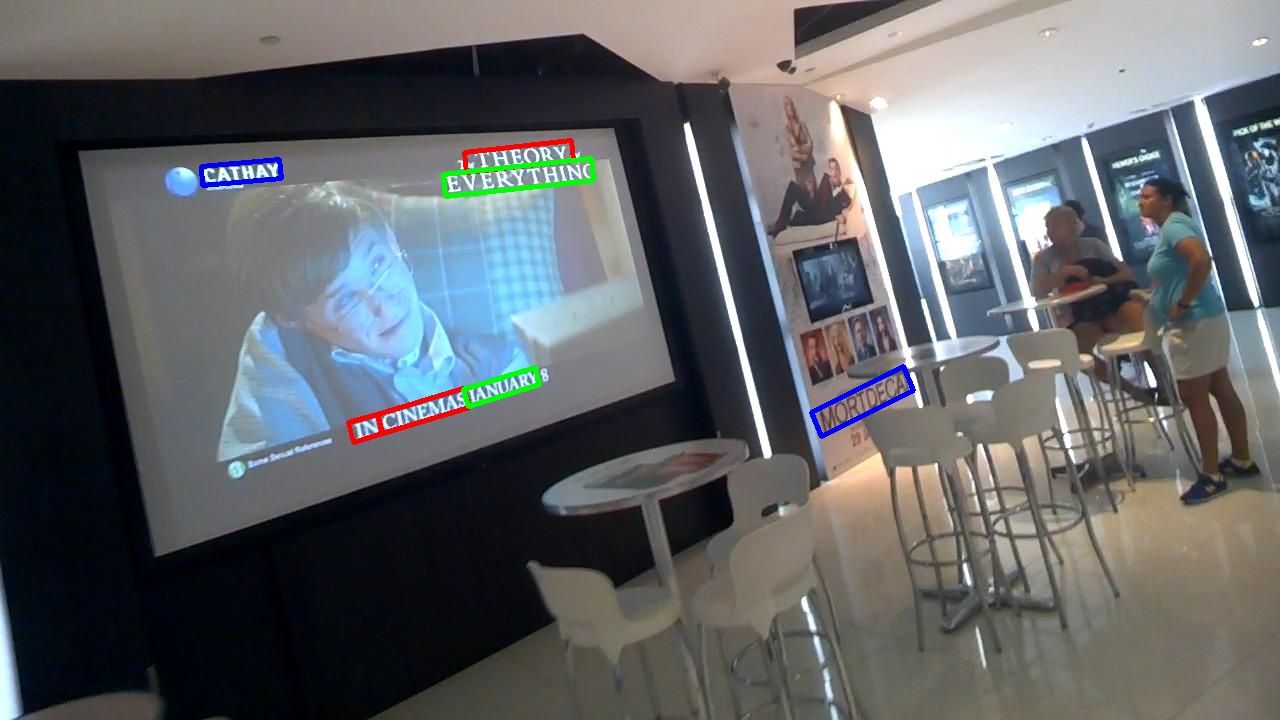}
  	\includegraphics[height=1.06in]{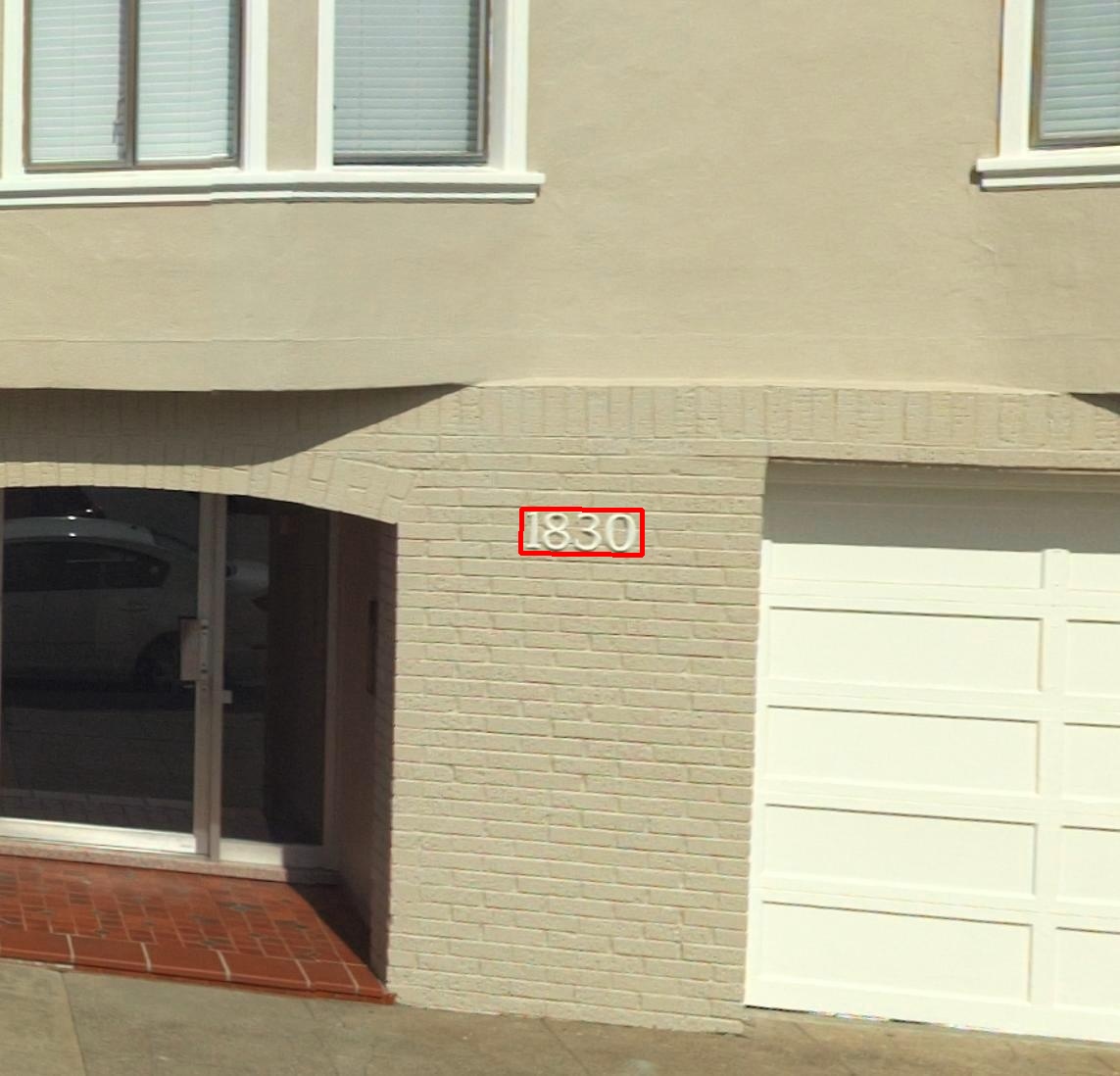}
  	\includegraphics[height=1.06in]{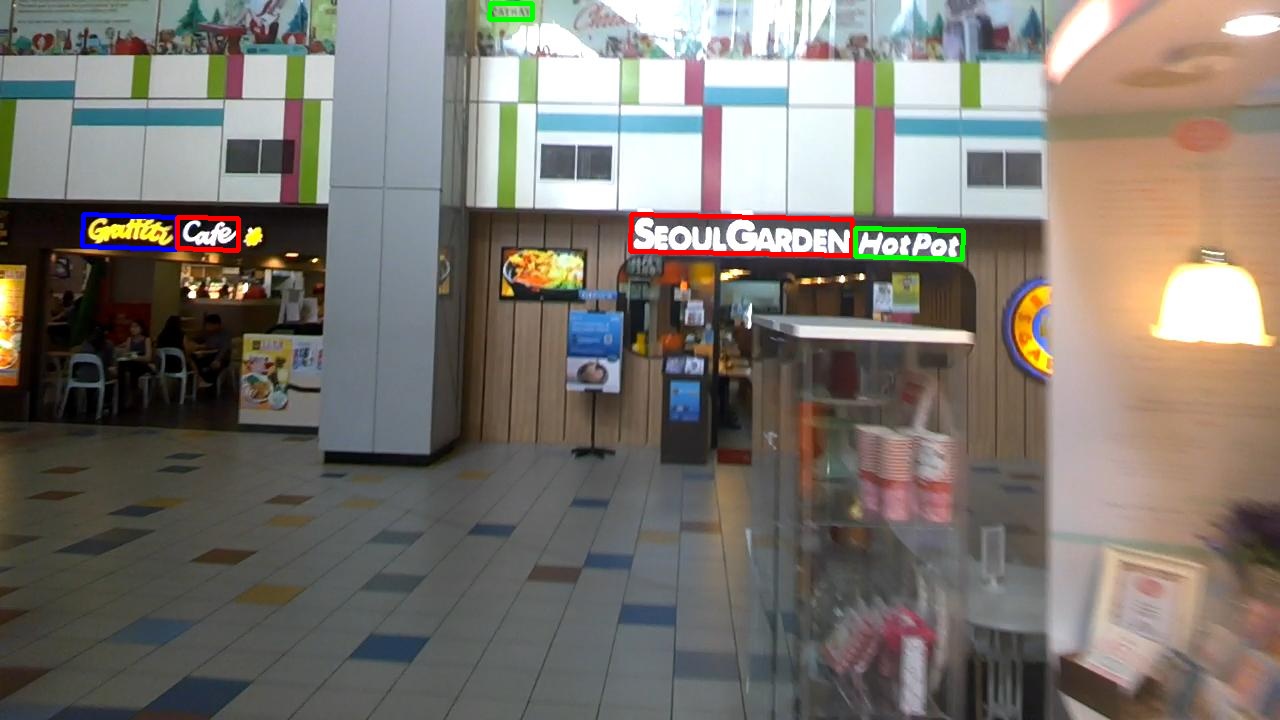}
  	\includegraphics[height=1.06in]{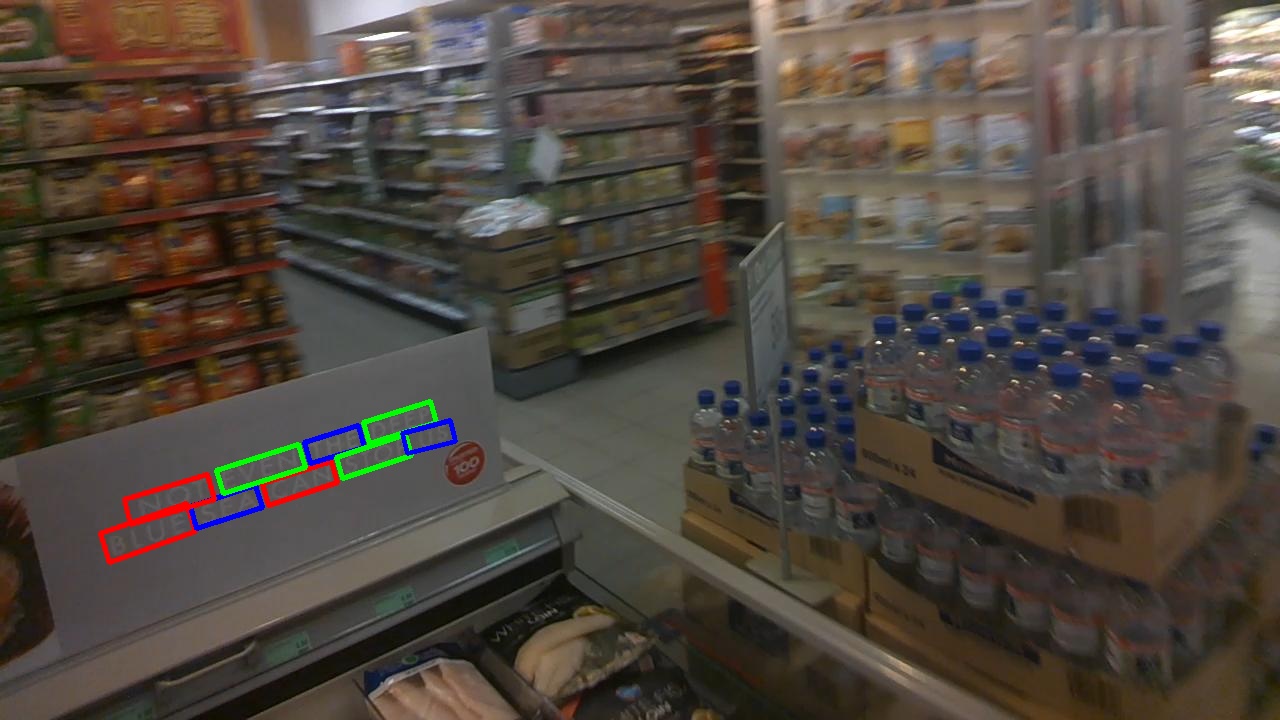}
  	
  	\includegraphics[height=1.06in]{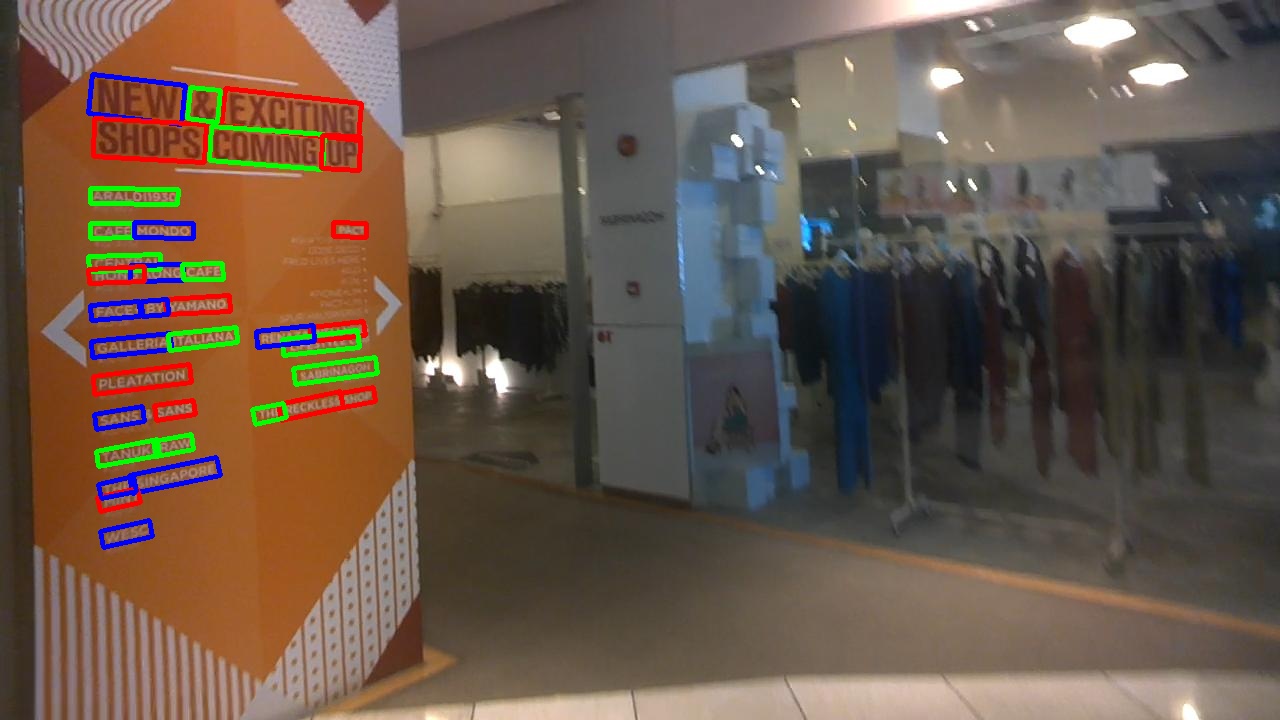}	
 	\includegraphics[height=1.06in]{final_result/ic15_img_491.jpg}
 	\includegraphics[height=1.06in]{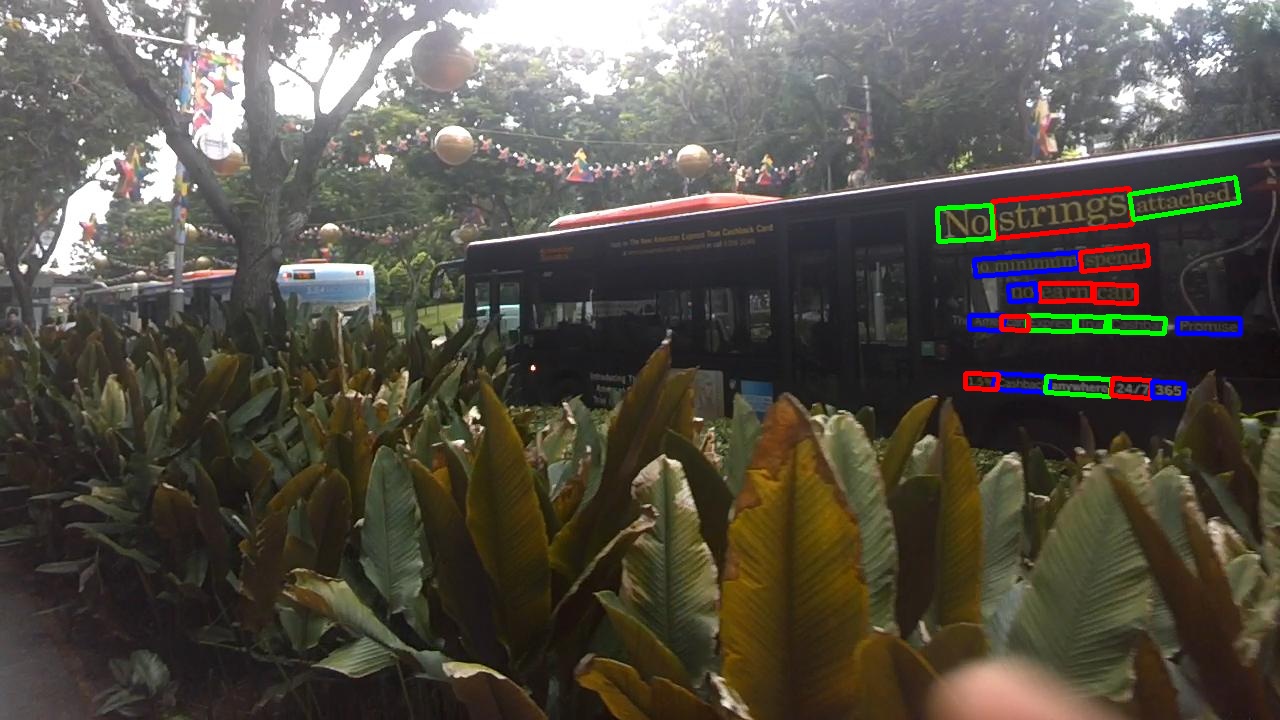}
 	\includegraphics[height=1.06in]{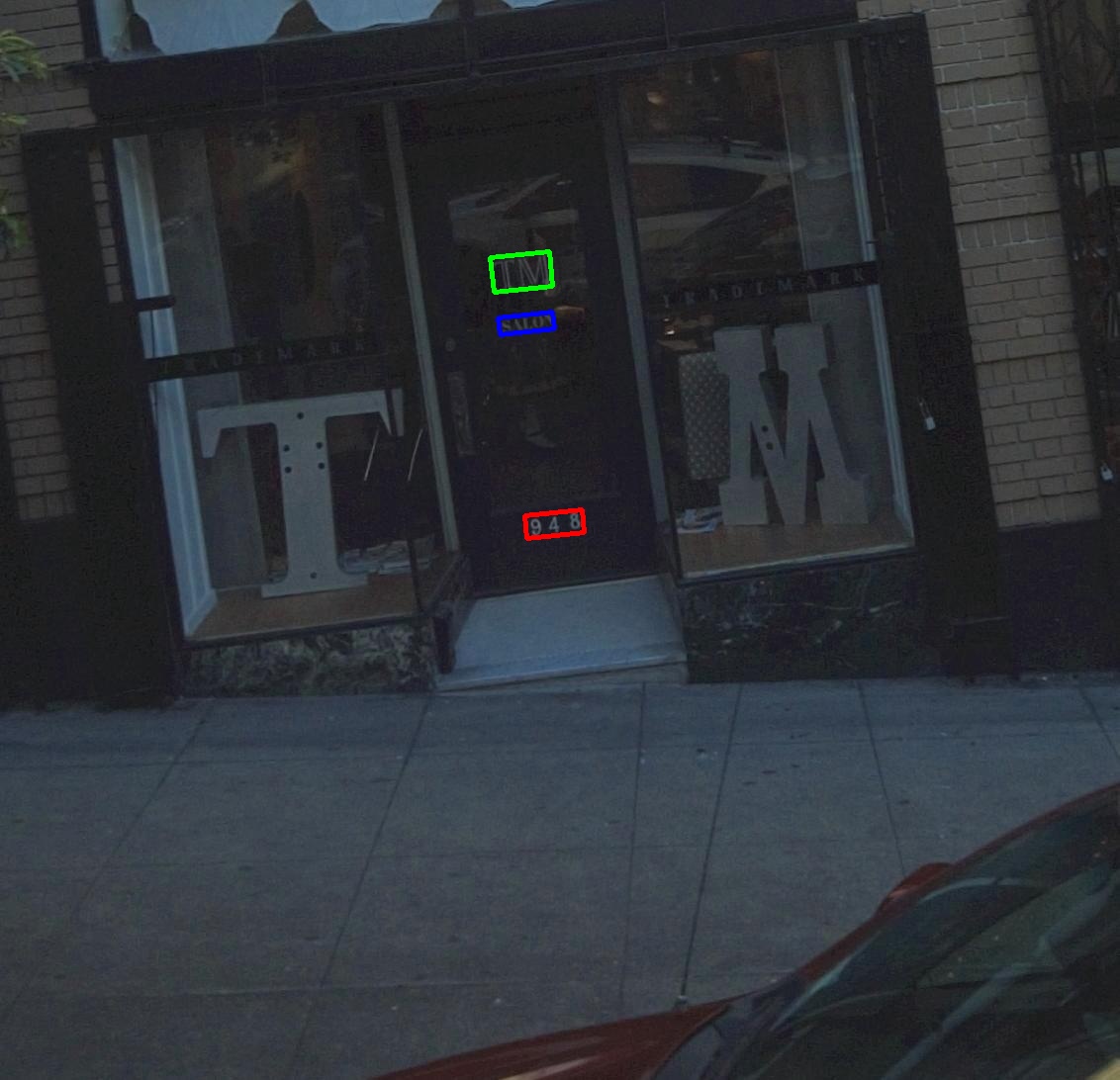}
 	\caption{Scene texts that have been successfully detected with the model trained with proposed framework. Images are from ICDAR2013, ICDAR2015, UberText dataset. Better when zoom in.}
 	\label{fig:FI_RES}
\end{figure*}

Following recent work~\cite{deng2018pixellink}, we categorize the scene text detection methods into three classes:
(1) Proposal based scene text detection~\cite{jiang2017r2cnn,liao2018textboxes++,shi2017detecting} which uses state-of-the-art object detection methods~\cite{ren2015faster,liu2016ssd} to classify and regress from each default proposal locations.
(2) Regression based scene text detection~\cite{he2017deep,Zhou_2017_CVPR} which usually does direct regression on each output pixel location as opposed to relative regression from default bounding boxes.
A text-nontext classification map is also needed to remove background nontext areas.
(3) Segmentation based scene text detection~\cite{He_2017_CVPR,zhang2016multi} which extracts text block at the first stage and uses low-level methods to obtain each individual text instances.

Due to the complex post-processing as well as the errors that incur from segmentation based methods, most recent methods follow direct regression based designs or proposal based designs.
Both of them achieved breakthroughs in multi-oriented scene text detection, and state-of-the-art end-to-end performance has been achieved by combining them with a scene text recognizer~\cite{Busta_2017_ICCV,Li_2017_ICCV}.

However, both of these categories have their drawbacks.
(1) Proposal based methods are usually less accurate in terms of recovering multi-oriented bounding boxes~\cite{he2017deep}.
(2) Regression based methods, though being able to recover accurately oriented bounding boxes, exhibit difficulties when the text has large variances of scales since each output bounding box is generated from a single output pixel.
Usually then, a multi-scale testing is needed.

Recently, He~\cite{he2017single} proposed to jointly learn a text attention map that suppresses background interference for a better word-level text proposal.
The text attention map is essentially a text-nontext segmentation, and could be used as a mask to remove background noise.
This design could be seen as a semantic segmentation based method that is used to assist text proposal generation.
Such a combination also achieved better performance.

In other research, learned contour detection~\cite{bertasius2015deepedge,yang2016object,liu2017richer} has become popular.
Unlike regular edge detection, learned contour information provides more instance-level semantic information.

Inspired by these two works, we also adopt the idea of using an easier task to assist scene text detection task.
However, instead of using a semantic text-nontext segmentation mask to suppress background interference, we propose to learn a contour map which directly encodes the text instance information to better assist the text detection task.

In summary, our contributions are:

(1) We introduce instance-level contour segmentation for scene text.
For general contour detection, contour could be seen as a subset of edges which preserve instance-level semantic meanings.
For scene text, instead of accurate edges for each letter, we propose that the contour of a word is the polygon that best encapsulates that word.

(2) We experimentally demonstrate that such an instance-level contour could be easily learned with a traditional encoder-decoder CNN design. Unlike contours in natural images which requires substantially more labor for annotation, our scene text contour needs only the original bounding box annotation.

(3) We propose to use the learned contour segmentation to assist scene text detection.
The contour map provides extra instance level semantic information which better assists the detection task compared to the instance-agnostic semantic information provided by the text nontext attention map~\cite{he2017single}.
It is also easier to learn than the regular text detection task which means that we follow the general design principle of using an easier task to assist a harder task.

(4) We propose two general designs for incorporating the contour task: contour as auxiliary loss, and contour as a multi-task cascade.
Extensive experiments on public dataset with different model configurations under the two frameworks has been conducted and show that both framework designs improve the performance of a state-of-the-art scene text detector.
Such improvements could not be achieved by simply choosing a deeper network backbone.

In the following sections, we discuss related works in scene text detection as well as contour segmentation.
In the section on \textit{Text Contour}, we discuss the contour definition as well as the basic CNN encoder-decoder network for contour detection.
In the section on \textit{Contour For Text Detection}, we introduce the framework for incorporating learned contour in scene text detection.
We then show the effectiveness of the proposed framework in the \textit{Experiments} Section.
Fig~\ref{fig:FI_RES} shows the detection results from a model trained using our framework.

\section{Related Work}
\subsection{Scene Text Detection}
Scene Text Detection has been a popular topic in computer vision community for a long time.
There have been naive sliding window methods~\cite{lee2011adaboost} with handcrafted features to low-level proposal based methods~\cite{epshtein2010detecting,huang2014robust,he2016aggregating}.
These early methods could only detect horizontal scene text.

Recent methods adopt deep learning to a greater degree with what is generally believed to be breakthrough results for multi-oriented scene text detection.
Jiang~\cite{jiang2017r2cnn} proposed to adapt the design of faster rcnn~\cite{ren2015faster} for quadrilateral word bounding box localization.
For each anchor box, it predicts the two corners as well as the short side length of the oriented box.
It belongs to the category of proposal based text detector for which the final boxes are regressed from default anchors.
He~\cite{he2017single} also adopted the text proposal based scheme but proposed to use a text-nontext attention map as feature map mask that removed unrelated background interference.

Zhou~\cite{Zhou_2017_CVPR} proposed a direct regression based method which is both robust and efficient.
It uses a traditional encoder-decoder design with each output location predicting both the text-nontext classification as well as the angle and distance to the four border of the bounding box.
Such a direct regression based framework has also been adopted by He~\cite{he2017deep} and both achieved good performance in oriented scene text localization.

\subsection{Contour Detection}
Contour detection was originally designed for extracting edges in natural images which preserve semantic meanings.
Usually additional effort is needed in labeling~\cite{yang2016object,liu2017richer} the images in order to train a learning based model.
For example, Yang~\cite{yang2016object} proposed to use a dense CRF to refine the annotation so as to obtain trainable contour annotations.
They also used the learned contour for proposal generation.
However, here we take a step further and use the learned contour directly as a feature to facilitate scene text detection which follows general multi-task learning design~\cite{caruana1997multitask}.

The text contour method defined in this work is unique in that:
(1) the contour is not a subset of low-level image edges but the border of each text instance,
(2) no extra annotation is needed which makes the current framework much easier to implement.

\section{Text Contour}
\label{sec:contour}
\subsection{Definition}
\begin{figure}[!htb]
    \centering
 	\includegraphics[height=0.55in]{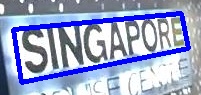}
 	\includegraphics[height=0.55in]{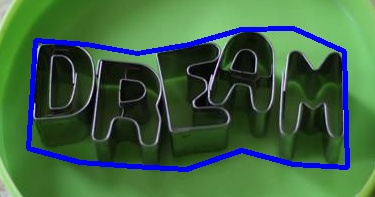}
 	\includegraphics[height=0.55in]{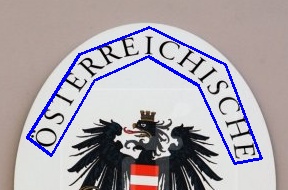}
 	\caption{Contour examples which are generated by polygon annotations. Most datasets provide quadrilateral annotations(e.g: ICDAR, UberText). The blue color is only for visualization.}
 	\label{fig:EG_CONTOUR}
\end{figure}

As opposed to the definition of contour in most of the previous literature, He~\cite{he2017multi} proposed to learn the contour of text, tables, and figures in PDF documents.
The contour does not align with any low-level image edges.
Instead, the network has to learn more semantic information in order to decide where is the contour.
This makes it an unique learning task.
Here we adopt a similar idea and define the contour for scene text to be the polygon that approximates the border of each instance.
The polygon could be a quadrilateral but also an octagon depending on the shape that we wish to learn and the annotation that the dataset could provide.
Fig~\ref{fig:EG_CONTOUR} provides several visualized examples of word contours.
Depending on the dataset annotation, different contour definitions could be used.

It is easy to notice that if we choose a quadrilateral as our setting, the contour could be rendered easily with the annotation provided by most datasets(e.g: ICDAR2015).
In this work we use a quadrilateral as our setting and demonstrate that the trained model could be helpful for text detection task.

\subsection{Learning to Extract Contour}
In order to learn to segment the contour of text, we design an encoder-decoder CNN with a skip link for multi-scale feature learning.
This will be the basic architecture for our contour network.
Sigmoid is applied to the last convolutional layer and we use a mean squared error(MSE) for training.
The loss value for training is denoted as $L_{contour}$.

\begin{figure}[!htb]
    \centering
 	\includegraphics[height=0.55in]{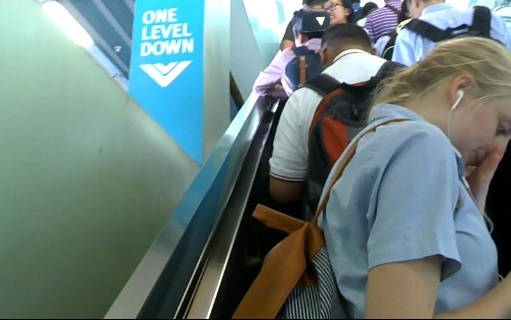}
  	\includegraphics[height=0.55in]{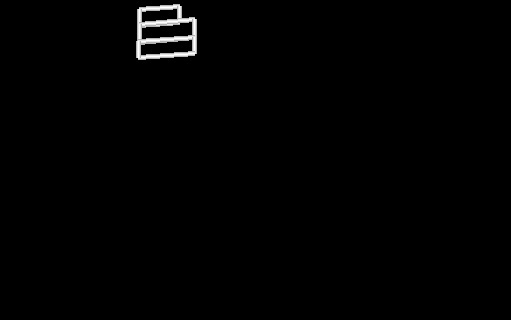}
  	\includegraphics[height=0.55in]{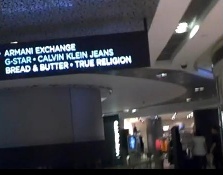}
  	\includegraphics[height=0.55in]{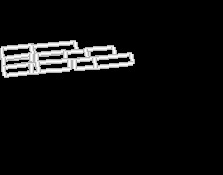}
  	
 	\includegraphics[height=0.7in]{gt_contour/1_13.jpg}
  	\includegraphics[height=0.7in]{gt_contour/1_13_contour.jpg}
  	\includegraphics[height=0.7in]{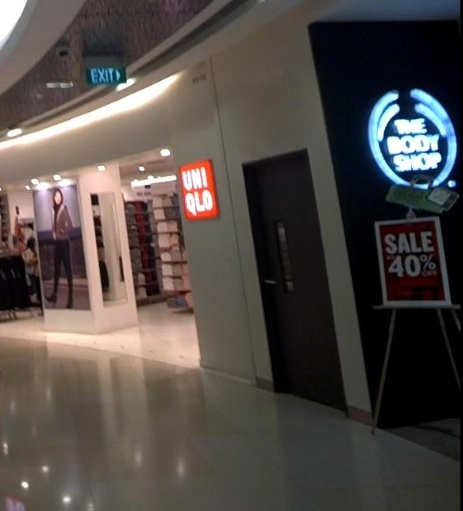}
  	\includegraphics[height=0.7in]{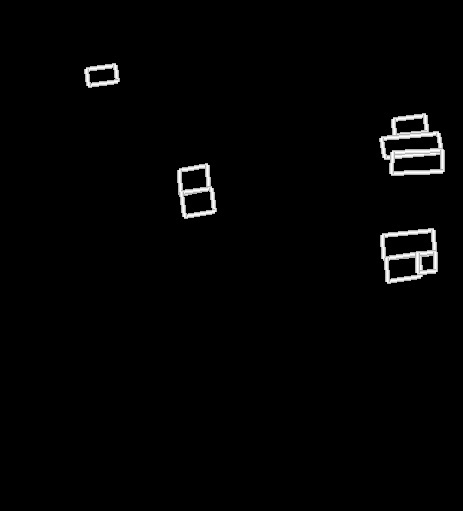}
 	\caption{The generated ground truth for training the contour network. Better when zoomed in.}
 	\label{fig:GT_CONTOUR}
\end{figure}

\textbf{Ground Truth Generation}
In order to train the network to detect scene text contour, we need to provide ground truths.
We use a quadrilateral bounding box to generate contour groundtruths.
Even though this might not capture the most precise boundary of scene text, a model trained with it can still give good prediction results.
Because of the imperfectness of the ground truth quadrilateral bounding boxes (as they are annotated for detection task), we follow the work of He~\cite{he2017multi} for creating a smoothed border for training.
Formally, we follow the equation~\ref{Eq:GT_CT} for generating the contour.
$S_{\mbox{contour}}$ represents the pixels in the contour generated based on the ground truth annotation.
Values of the regression targets are empirically selected.
This way a contour prediction that is a few pixels away from the ground truth will not be penalized that much.

\begin{equation} \label{Eq:GT_CT} 
	x_i = 
	\begin{cases}
		1     & \quad \text{if } i \in S_{\mbox{contour}} \\
	    0.9   & \quad \text{if dist}(i,j) == 1 \text{ and } \exists j \in S_{\mbox{contour}} \\
	    0.6   & \quad \text{if dist}(i,j) <= 3 \text{ and } \exists j \in S_{\mbox{contour}} \\
	\end{cases}
\end{equation}
Fig~\ref{fig:GT_CONTOUR} shows an example input image and the corresponding contour ground truth.

\begin{figure}[!htb]
    \centering
 	\includegraphics[height=1.2in]{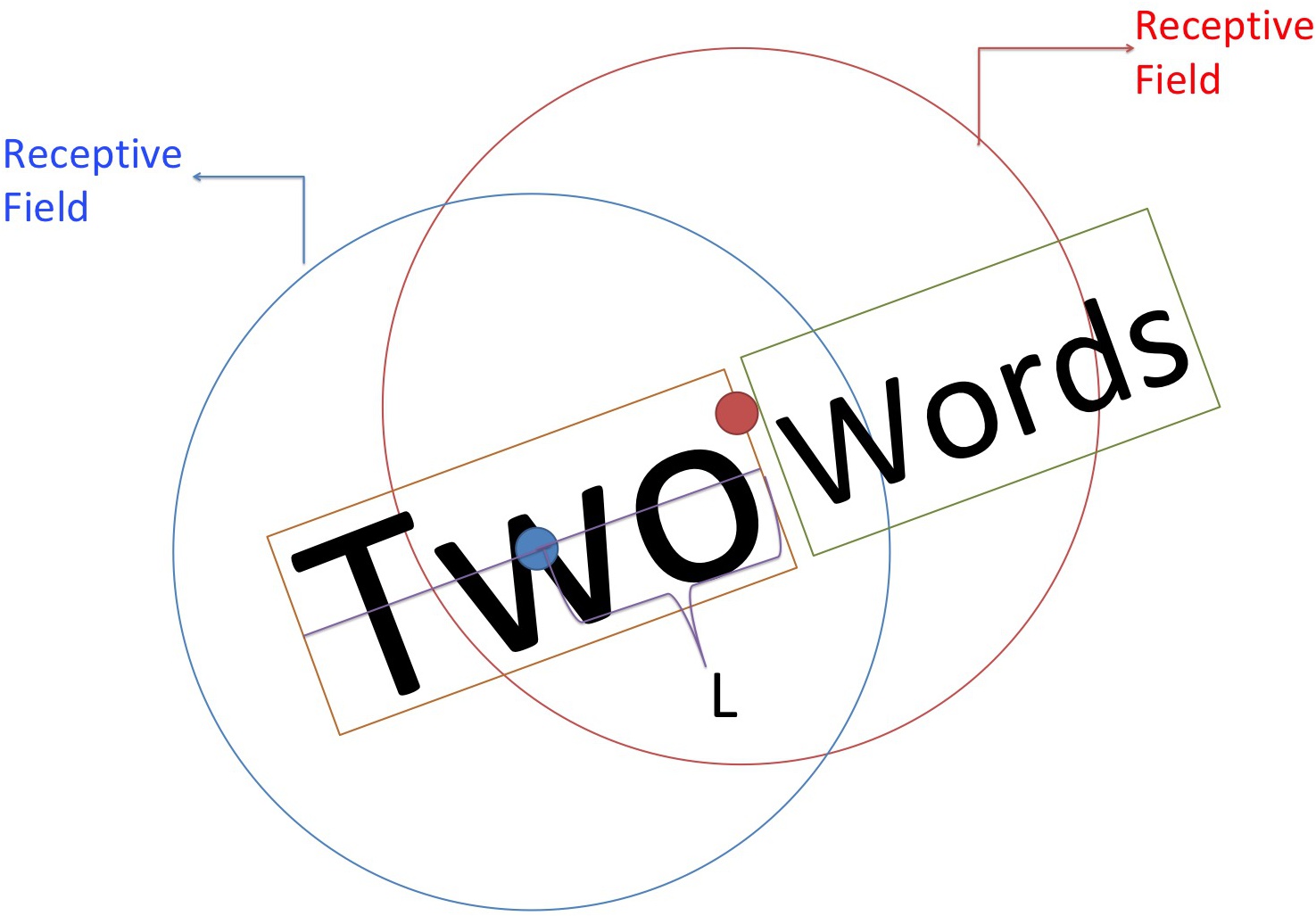}
 	\caption{Illustration of the reason that contour detection is easy to learn and can help scene text detection.}
 	\label{fig:LEARNABILITY}
\end{figure}

\textbf{Learnability}
Even though the contour segmentation task needs to learn instance-level information, we claim that the task is easier to learn when compared with learning a regular detection task.
The major reason is that the network only needs to learn which pixels separate text from non-text background or separate two text instances (if they are close).
It implicitly learns instance information without the need for global context.
This is because even when looking at a relatively local region, it is still possible to identify whether it is the boundary or not.
This is opposed to the regression task in scene text detection, for which one needs to identify the distance to text instance border or corner.
Fig.~\ref{fig:LEARNABILITY} illustrates this idea.
Suppose that the blue point and the red point represent the output pixels for a text detection network and a contour detection network, respectively.
The blue and red circle represents their receptive fields.
$L$ represents one of the regression targets (The distance to a border of text).
The text detector has to identity where is the border of the text instance and how far the current pixel is from the boundary.
This is much harder when centered at the boundary and needs to predict the distance to the other side boundary of the text bounding box.
However, predicting whether the pixel belongs to a border(contour) is much easier.
On the other hand, if we could provide rough contour information to the network, the regression task can be much easier to learn.
This is the main idea why we use an easier task to assist scene text detection.

\section{Contour For Text Detection}
\label{seg:contour_for_det}   
Since we are able to learn the text contour from input image, we propose to use it as an additional task for scene text detection.
We describe two proposed frameworks for using text contours for scene text detection as well as the scene text detector that we adopted in this work.
Basically, our two general frameworks are:
(1) Use the contour segmentation as a sub-task and jointly train the network.
This we call Auxiliary TextContourNet.
(2) Use the learned contour as features for scene text detection in a cascade fashion.
This we call Cascade TextContourNet.

\subsection{Auxiliary TextContourNet}
\begin{figure}[!htb]
    \centering
 	\includegraphics[height=0.95in]{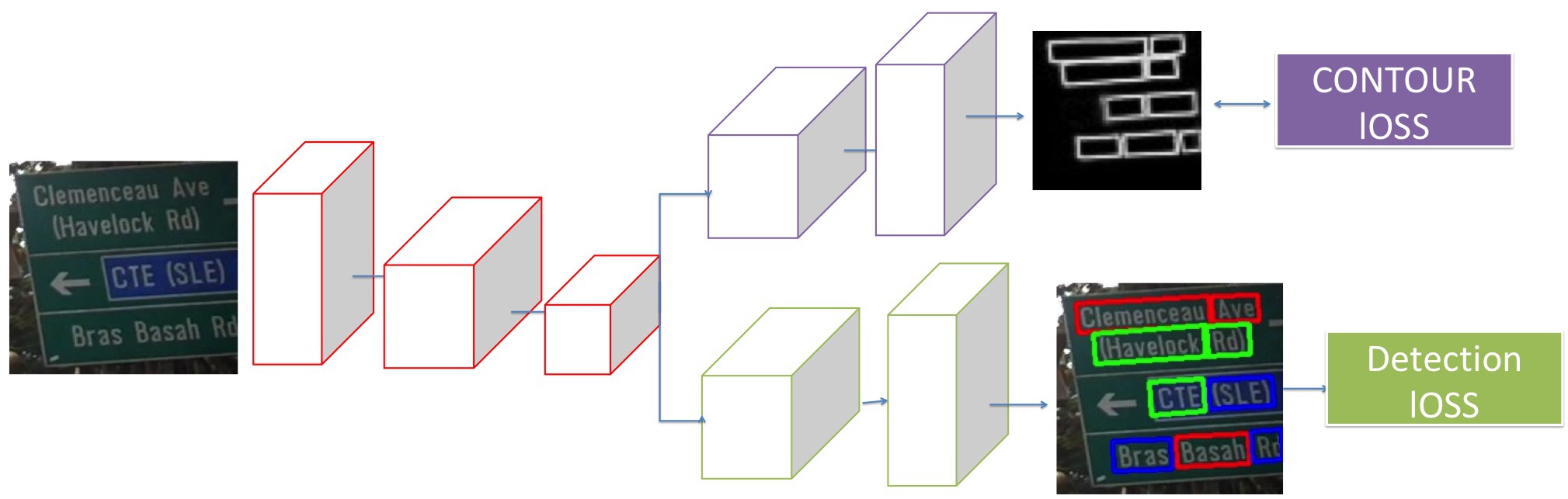}
 	\caption{Auxiliary TextContourNet: A general framework of adding the contour task as an auxiliary loss.}
 	\label{fig:auxiliary_pipeline}
\end{figure}

The framework for Auxiliary TextContourNet is shown in Fig.~\ref{fig:auxiliary_pipeline}.
It follows basic encoder-decoder design with several layers of shared convolutions.
Note that in the illustration of the framework, only the encoder convolution layers are shared due to space limitations.
In our experiments, we show results of two designs with different convolutional layers shared.

\subsection{Cascade TextContourNet}
The contour could also be used in a cascade fashion.
As opposed to previous work~\cite{he2017single} which applies the semantic text-nontext segmentation map as a mask, this framework treats the explicitly learned contour as another feature layer and lets the network jointly learn from both the visual features extracted from original image as well as the contour prediction map.

We believe this framework has several advantages:
(1) Applying the semantic text-nontext map as mask makes it hard for the network to recover errors from the segmentation task.
Instead, here we jointly learn the task from contours as well as the extracted visual features.
Errors made from contour network doesn't necessarily lead to false detection results.
(2) The text-nontext prediction is a regular task for many scene text detector~\cite{Zhou_2017_CVPR,he2017deep,deng2018pixellink}.
Adopting it as a subtask will have no benefit for these detector.
(3) The text-nontext segmentation only provides semantic information while contour segmentation provides instance-level semantic information.
Being able to provide instance-level information in the early stage of the network allows the network to learn to propose an instance bounding box more easily and accurately.

This also follows our general design intuition of using an easier task to assist the harder task.
Previous works~\cite{jaderberg2016reinforcement,adi2016fine} also use such ideas for training deep models.
Such a cascade fashion has also been adopted by other works~\cite{dai2016instance,zhang2016joint}.

In our task, both settings could be adopted and we experimentally show that both of them improve the performance of the original scene text detector.
By explicitly learning to segment the contour and using it as features for detection(\textit{Cascade TextContourNet}), the detector could perform better than using the contour detection as an auxiliary loss(\textit{Auxiliary TextContourNet}).

\begin{figure}[!htb]
    \centering
 	\includegraphics[height=0.95in]{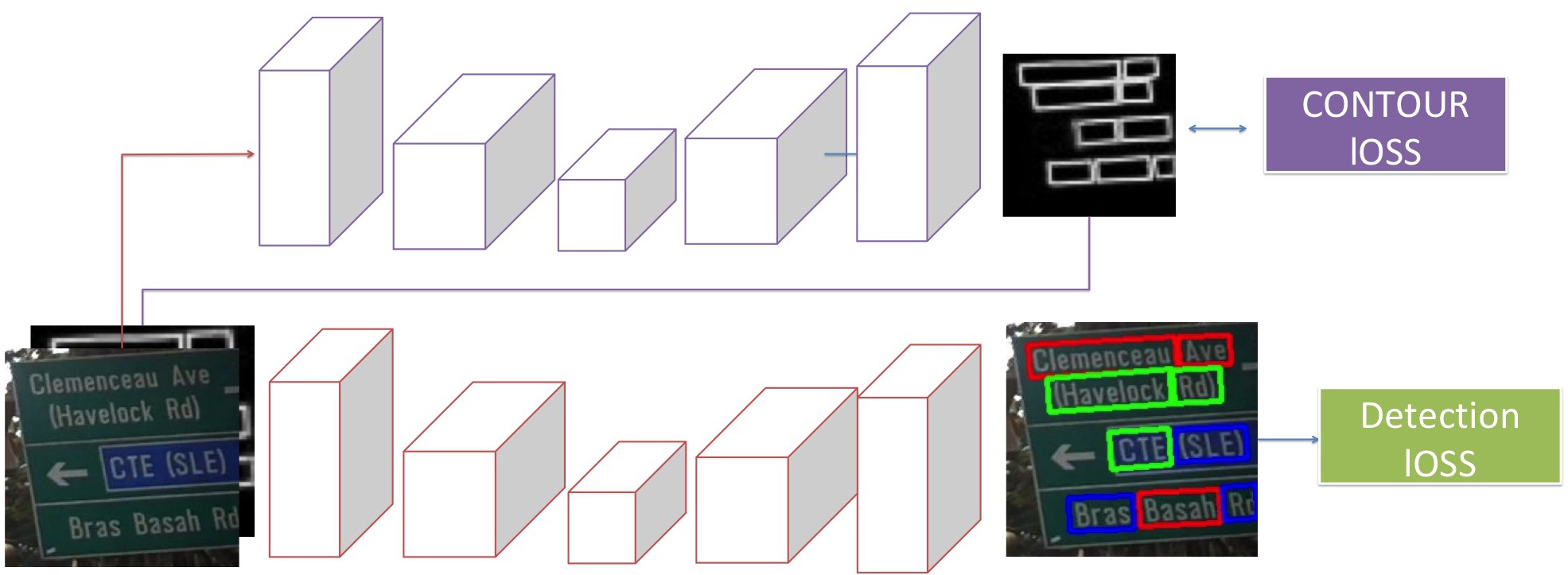}
 	
 	\includegraphics[height=0.95in]{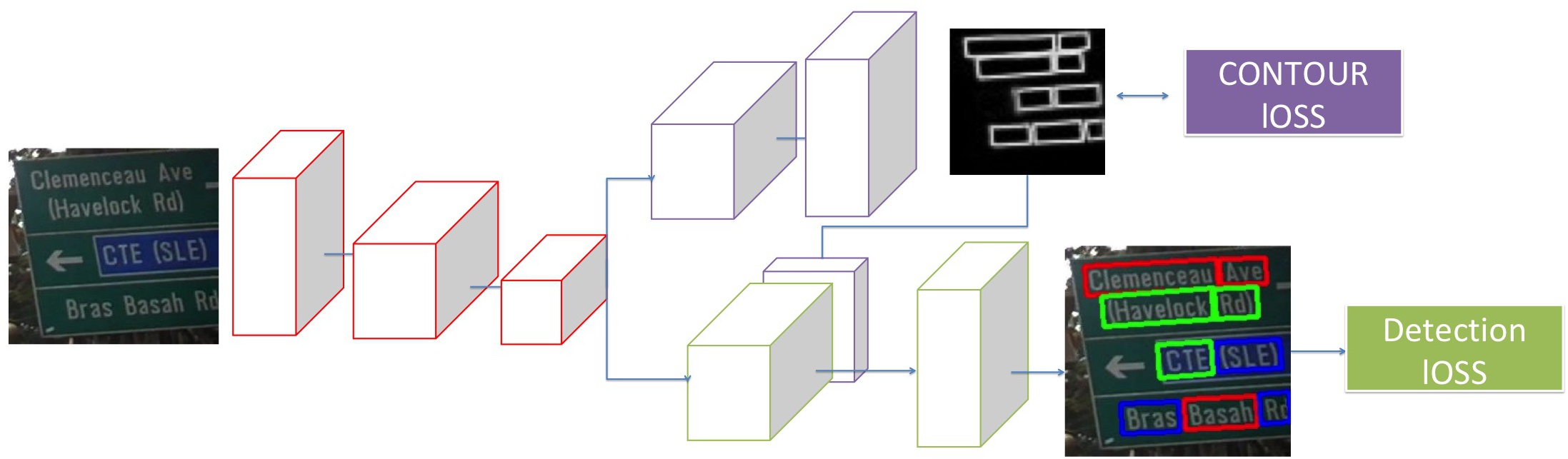}
 	\caption{Two frameworks for using in a cascade manner the contour information for scene text detection. \textbf{up}: Join the contour in the early stage. In this case, there is no shared convolutional parameters. \textbf{down}: Join the contour in the later stage and share the features and parameters in the encoder.}
 	\label{fig:MergeStage}
\end{figure}

In order to use the contour segmentation output as features, we propose two schemes: (1) Early Merge: merge the extracted contour information in the early encoder stage.  (2) Late Merge: Merge the extracted contour information in later decoder stage.
Fig~\ref{fig:MergeStage} shows the visualization of the proposed frameworks.
Note that, for early merge, since we have to recompute the encoder features and the input to the detection network is a 4-channel tensor, the convolutional features are not shared for the detection task and the contour segmentation task.

\subsection{Scene Text Detector}
We adopt the work~\cite{Zhou_2017_CVPR} as our base detector design.
It is one of the representative, state-of-the direct regression based scene text detector models.
Here we give a brief description of the method.

Given an image $I$ with height $I_h$ and width $I_w$, an encoder-decoder network is used to produce $M$ channels of output with height and width $S_h$, $S_w$, respectively.
The value of $M$ depends on the different geometric shapes the model is trained to predict.
Two geometries were proposed in the original paper~\cite{Zhou_2017_CVPR} and here we use $RBOX$ setting.

In $RBOX$, a rotated bounding box is predicted for each output pixel location.
There are 6 output maps in total and the first channel is the score map with each pixel valued from [0,1].
It corresponds to the confidence for each location to be text or not.
Note that this is also a major difference between a direct regression based method and a proposal based method.
For a proposal based method, the class of each output location depends on the intersection over the union(IOU) score of the default anchor and the ground truth bounding boxes.
Here the class simply depends on whether the pixel belongs to the region of text or not.
It should be noted that the ground truth bounding boxes are shrunk before creating the score map targets to better separate out nearby words.
Dice loss is used for training the score map, which directly optimizes the IOU of the segmentation results.
We denothe this loss as $L_{score}$ as in~\ref{eq:dice_loss}.
$y_s^*$ denotes the ground truth score map and $\hat{y}_s$ is the predicted score map.
$\beta$ is the weight for positive class and negative class.

\begin{align}
	L_{score}=1 - \frac{2 \times \hat{y}_s * y^*}{(\sum{\hat{y}_s} + \sum{y^*})} \label{eq:dice_loss}
\end{align}

The other output channels correspond to the geometric information of the predicted bounding boxes.
For the $RBOX$ scheme for each positive class location, its distances to the 4 boundaries of the rotated bounding box are used as ground truth, and IOU loss~\cite{Yu:2016:UAO:2964284.2967274} is used for loss calculation because it's invariant to object scale such that different scales of text will have the same contribution.
The loss is denoted as $L_{IOU}$.
The orientation angle of the word $\hat{\theta}$ is also used as another target and the loss is denoted as~$L_\theta(\hat{\theta}, \theta^*)$, where $\theta^*$ is the ground truth angle.

As such, 5 channels for geometry will be predicted with the total geometry loss denoted as $L_{geo}$.
The total training loss $L_{det}$ is the weighted sum of $L_{score}$ and $L_{geo}$ in Eq.~\ref{eq:seg_det_loss}.
More details can be found in the original paper~\cite{Zhou_2017_CVPR}.

\begin{align}
\begin{cases}
	L_\theta(\hat{\theta}, \theta^*) = 1 - \cos(\hat{\theta} - \theta^*) \\
	L_{geo} = \lambda_{IOU} L_{IOU} + L_\theta(\hat{\theta}, \theta^*) \\
\end{cases}
\end{align}  

\begin{align}
L_{det} = \lambda_{geo} + \lambda_{cls} L_{score} \label{eq:seg_det_loss}
\end{align} 

\subsection{Joint Training}
By incorporating the contour detection with scene text detection, the loss for both frameworks is defined in~\ref{eq:total_loss}.
$\beta$ is set to 0.1 and $\lambda_{cls}$ is set to 0.01 for our experiments.

\begin{align}
	L = L_{geo} + \lambda_{cls} L_{score} + \beta L_{contour} \label{eq:total_loss}
\end{align} 

\section{Experiments}
\label{sec:Experiments}
\subsection{Implementation}
The pipeline is implemented in the Tensorflow~\cite{abadi2016tensorflow} framework.
The baseline method is based on the East implementation~\footnote{https://github.com/argman/EAST} which is a modified version of the original model~\cite{Zhou_2017_CVPR} with slightly better performance.
Data augmentation during training was similar to SSD~\cite{liu2016ssd} with random cropping and scaling.
The training process contains two steps:
(1)train all the augmented images with fixed input size 512$\times$512. (2) fine tune the trained model with fixed input size 768$\times$768.
We use Adam optimizer for training.

All models have a CNN backbone of Resnet50~\cite{he2016deep} with a feature pyramid design~\cite{lin2017feature} unless specified otherwise.
The CNN is initialized with pretrained image classification model.
The output resolution is 1/4 of the input image resolution.
The CNN backbone part is initialized with a pretrained model.
For the Auxiliary TextContourNet framework, we implement two variants of it:
(1) Only the CNN encoder(Resnet50) parts are shared($Auxiliary ContourNet_1$).
(2) The CNN encoder and decoder parts are all shared($Auxiliary ContourNet_2$).
The only difference between $Auxiliary ContourNet_2$ with the baseline model is that an additional output channel is produced and trained as the contour segmentation.
For Cascade TextContourNet, we designed two models corresponding to the two scheme proposed:
(1) Early Merge: we resize and concatenate the output contour to feed as one input channel($Cascade ContourNet_1$).
The input to the detection network is thus a 4-channel tensor.
(2) Late Merge: we concatenate the output contour with the last layer in the detector branch in depth dimension. Then three convolutional layers are added with depth 32, kernel size 3$\times$3 before producing the final detection output($Cascade ContourNet_2$).

\subsection{Dataset Description}
In order to demonstrate the effectiveness of the proposed framework, we investigated various datasets discussed next.


\subsubsection{ICDAR 2013}
\textit{Focused Scene Text Detection Dataset}.
It contains 229 and 233 training and testing images, respectively. All the text are horizontal or close to horizontal.
Horizontal rectangles are annotated for each word in the image.
The quadrilateral is simply generated from the horizontal bounding box.
We denote the training set and testing set as $IC13_{train}$ and $IC13_{test}$.

\subsubsection{ICDAR2015}
\textit{Incidental Scene Text Detection Dataset}.
This contains 1000 training images and 500 testing images denoted as $IC15_{train}$ and $IC15_{test}$.
Images are taken with portable devices with motion blur.
The text are multi-oriented with quadrilateral annotation.

Both ICDAR2015 and ICDAR2013 dataset are widely used with relatively more accurate quadrilateral annotation.
In our experimental setting, they are used to demonstrate that with an accurate quadrilateral annotation, our model can improve the performance of the baseline.

\subsubsection{UberText}
UberText~\cite{zhang2017uber} dataset is a newly released dataset with images taken from street views.
It contains both 1K and 4K resolution images with training, testing, and validation splits.
We use the 1K version in our experiment.
We only use the training and testing splits for this evaluation.
It contains 16927 training images and 10157 testing images.
To the best of our knowledge, it is currently the largest scene text dataset with quadrilateral bounding box annotation.
However, it also has its minor drawbacks.
A few annotations are not consistent and accurate.
The training set and testing set are denoted as $Uber_{train}$ and $Uber_{test}$, respectively.

\subsection{Quantitative Experiments}
\subsubsection{Capacity Study}
In this study, we demonstrate the capacity of the proposed framework by training on $IC13_{train} + IC13_{test} + IC15_{train} + IC15_{test}$ and test on $IC15_{test}$.
This experiment aims at showing that with quadrilateral annotation, the contour task can \textit{substantially} improve the quality of the feature representation learned and the model's ability in fitting the training data.
With potentially much more data that are commonly used in industry, such study shows that by adopting our framework, the model can learn a much better feature representation from the training data and to achieve a much better performance.
The results are in Table~\ref{table:IC15_det_res_train}.

We can see that adding contour as a auxiliary task can improve the f-measure by approximately 1 percent while adding it as a cascade task can improve it by 3 percent with the late merge mechanism（$Cascade ContourNet_2$).
We also observed that for early merge cascaded model（$Cascade ContourNet_1$） the performance dropped instead.
We believe that this is because the learned contour map is highly semantic and the network could not learn good features when we concatenate it with the raw input images.
In later experiments, we only use late merge for comparison.

Baseline 101 represents the model trained with Resnet101 backbone.
It is also initialized from a pretrained model.
We can see that even with a much deeper network, the capacity of the network almost stayed the same.

All the following studies correspond to a regular experimental setting.
They could be seen as evaluating the generalizability of the proposed framework.

\begin{table}
  \begin{tabular}{|c||l|l|l||l|l}
    \hline
    \multirow{2}{*}{Method} & \multicolumn{3}{c}{ICDAR2015}\\
    & Recall & Precision & F-1\\
    \hline
    Baseline & 86.2 & 90.40 & 88.25\\
    Baseline 101 & 86.72 & 90.95 & 88.78\\
    \hline
    $Auxiliary ContourNet_1$ &86.42 &92.43 &89.32\\
    $Auxiliary ContourNet_2$ &86.61 &92.88 &89.635\\
    $Cascade ContourNet_1$ &85.22 &84.33 &84.77 \\
    $Cascade ContourNet_2$ & \textbf{91.42} & \textbf{93.44} & \textbf{92.58}\\
  \end{tabular}
  \caption{Capacity study(Ability in fitting training data): Localization Performance(\%) on ICDAR2015 dataset. We use this study to show how well the model can fit the training data.}
  \label{table:IC15_det_res_train}
\end{table}



\subsubsection{ICDAR2013}
We train the model with different proposed framework configurations on $IC13_{train} + IC15_{train}$ and test on $IC13_{test}$
with results in Table ~\ref{table:IC13_det_res}.

\begin{table}
  \begin{tabular}{|c||l|l|l||l|}
    \hline
    \multirow{2}{*}{Method} &\multicolumn{3}{c}{ICDAR2013}\\
    & R & P & F-1 \\
    \hline
    Zhang et al~\cite{zhang2016multi} & 89 & 78 & 83\\
    StradVision~\cite{karatzas2015icdar} & 79 &90 & 84\\
    Multi-scale~\cite{He_2017_CVPR} & 79 & 83 & 85\\
    Yao et al~\cite{yao2016scene} & 80 & 89 & 84\\
    DirectReg~\cite{he2017deep} &\textbf{92} &81 &86\\
    \hline
    Baseline MS &83.61 &91.21 &87.24\\
    \hline
    $Auxiliary ContourNet_1$ MS &84.20 &92.48 &88.15\\
    $Auxiliary ContourNet_2$ MS &84.04 &\textbf{93.40} &88.47\\
    $Cascade ContourNet_2$ MS &85.22 &93.02 &\textbf{88.95}\\
    \hline
  \end{tabular}
  \caption{Localization Performance(\%) on ICDAR2013 dataset. R: Recall. P: Precision. \textbf{MS} represents multi-scale testing. We compare our model with the baseline as well as other methods.}
  \label{table:IC13_det_res}
\end{table}
 
\subsubsection{ICDAR2015}
We train the model with different proposed framework configurations on $IC13_{train} + IC15_{train}$ and test on $IC15_{test}$.
The trained model is the same one as tested for ICDAR2013 with results in Table~\ref{table:IC15_det_res}.
We can see that, By adopting our framework, we improve the model slightly and the cascade setting achieves the best performance.

\begin{table}
  \begin{tabular}{|c||l|l|l||l|l}
    \hline
    \multirow{2}{*}{Method} & \multicolumn{3}{c}{ICDAR2015}\\
    & R & P & F-1\\
    \hline
    Zhang et al~\cite{zhang2016multi} &43 & 71 & 54\\
    StradVision~\cite{karatzas2015icdar} &37 & 77 & 50\\
    Multi-scale~\cite{He_2017_CVPR} &54 &76 &63\\
    Yao et al~\cite{yao2016scene} & 59 & 72 & 65\\
    DirectReg~\cite{he2017deep} &82 &80 &81\\
    SegLink~\cite{shi2017detecting} &77 &73 &75\\
    \hline
    Baseline &77.90 &82.38 &80.07\\
    \hline
    $Auxiliary ContourNet_1$ &77.35 &83.88 &80.48 \\
    $Auxiliary ContourNet_2$ &78.58 &85.02 &81.64 \\
    $Cascade ContourNet_2$ &\textbf{79.91} &\textbf{86.12} &\textbf{82.90} \\
    \hline
  \end{tabular}
  \caption{Localization Performance(\%) on ICDAR2015 dataset. R: Recall. P: Precision.}
  \label{table:IC15_det_res}
\end{table}

\subsubsection{UberText}
We train on $Uber_{train}$ and test on $Uber_{test}$.
Uber text also contains quadrilateral annotation.
Here we simply fit a \textit{minimum oriented rectangle} for training and testing.
For other baseline approaches, we implemented the faster RCNN~\cite{ren2015faster} based on Google Object Detection api~\footnote{github.com/tensorflow/models/tree/master/research/object$\_$detection} and modified the regression scheme as in Equation~\ref{eq:pro_reg_transform}.

\begin{align}
	t^*_{xj} = (g_{xj} - x_a)/w_a  \nonumber \\
	t^*_{yj} = (g_{yj} - y_a)/h_a \nonumber \\
	j = 1, 2, 3, 4\label{eq:pro_reg_transform}
\end{align}

$t^*_{xj}$ and $t^*_{yj}$ is the encoded $x$, $y$ coordinates, respectively.
They follow the same design pattern as in the original faster rcnn for object detection when we encode the center.
Here each oriented bounding box, we need to encode 4 coordinates.
Note that here we ignore the proposal index $i$ that is used in the following equation.

By using such regression scheme, the overall loss function for training this baseline faster rcnn scene text detector is as~\ref{eq:baseline_loss}

\begin{multline}
	L_{baseline} = \frac{1}{N_{cls}}\sum_{i} L_{cls}(p_i, p_i^*) + \\
	\lambda *  \frac{1}{N_{reg}}\sum_{i} p_{i}^* \sum_{j=1}^{4} L_{reg}(t_{ij}, t_{ij}^*) \label{eq:baseline_loss}
\end{multline}

$L_{cls}$ represents the classification loss for each selected anchors.
We use cross entropy for it.
$L_{reg}$ represents the regression loss for each matched detected bounding boxes.
We use smooth $L_{1}$ loss for it.
The baseline method is called $TextFasterRCNN$.

We trained two faster-rcnn model with the backbones resnet50 and resnet101
with results  in~\ref{table:Uber_det_res}.

\begin{table}
  \begin{tabular}{|c||l|l|l||l|l|}
    \hline
    \multirow{2}{*}{Method} & \multicolumn{3}{c}{Uber Text}\\
    & Recall & Precision & F-1\\
    \hline 
    $TextFasterRCNN$ resnet101 &55.80 &53.04 &54.38\\
	$TextFasterRCNN$ resnet50 &49.74 &47.29 &48.48\\
    \hline
    Baseline & 70.33 & 83.82 &76.48\\
    \hline
    $Auxiliary ContourNet_1$ &73.57 &82.92 &77.96\\
    $Auxiliary ContourNet_2$ &73.33 &82.53 &77.64\\
    $Cascade ContourNet_2$ &\textbf{74.02}&\textbf{84.26}&\textbf{78.81}\\
    \hline
  \end{tabular}
  \caption{Localization Performance(\%) on UberText dataset. $TextFasterRCNN$ is a proposed baseline method based on tensorflow object detection API.}
  \label{table:Uber_det_res}
\end{table}

\subsection{Consistency}
\begin{figure}[!htb]
    \centering
 	\includegraphics[height=1.4in]{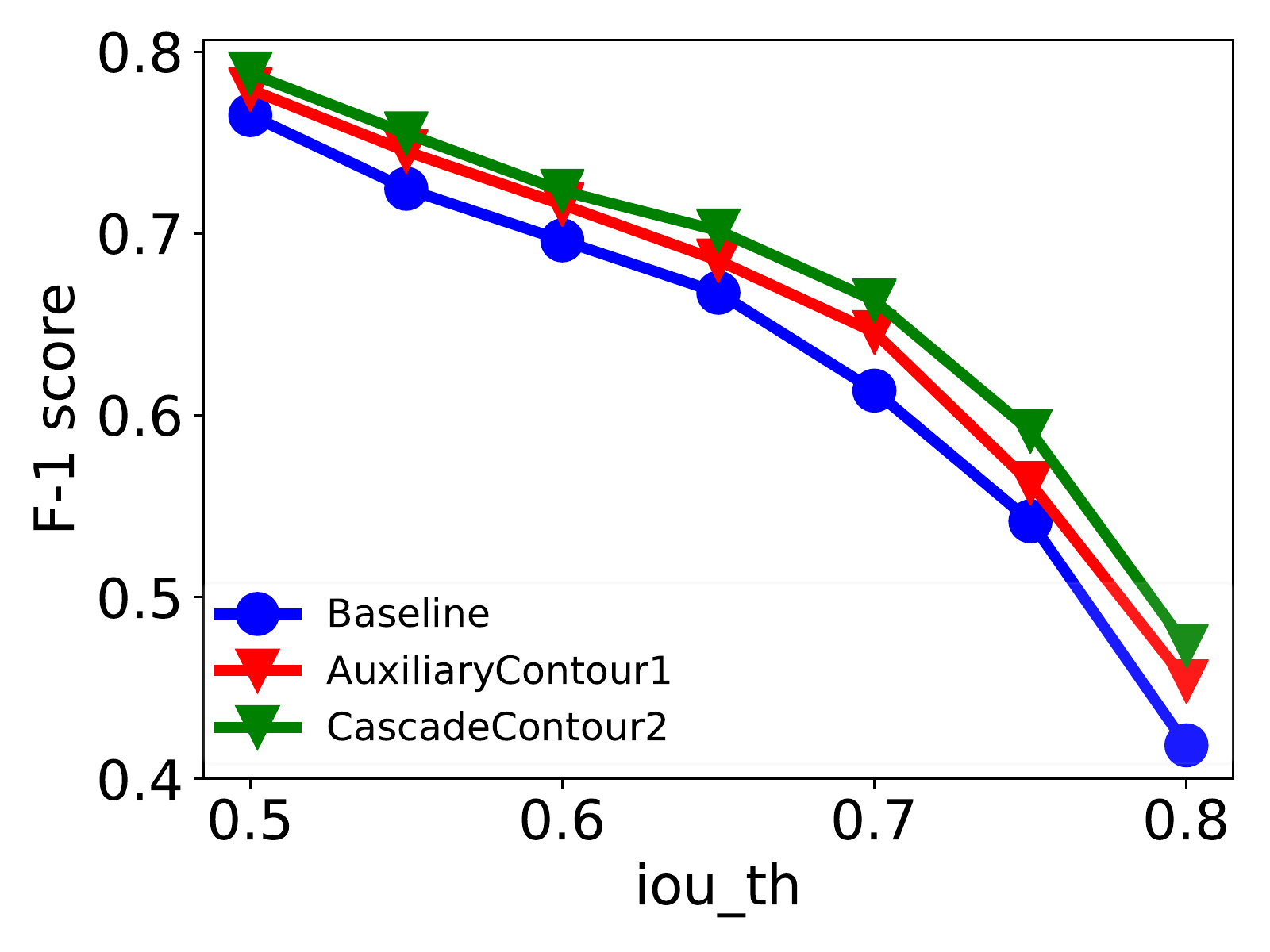}
 	\caption{F-1 score w.r.t the IOU threshold in scene text detection evaluation for three different models. The $Cascade ContourNet_2$ and $Auxiliary ContourNet_1$ consistently outperform baseline model.}
 	\label{fig:Dif_IOU}
\end{figure}

Scene text detection is usually evaluated and compared based on an quadrilateral IOU threshold 0.5.
However, such an IOU threshold may not reflect the model's real performance when combining it with a scene text reader.
This is because usually we need a much higher IOU in order to read the text correctly.
Such an idea is also mentioned in~\cite{deng2018pixellink}.
Here we show that the model trained with our framework could consistently outperform a baseline method with a different IOU threshold.
The results are in Fig.~\ref{fig:Dif_IOU}.
We only show the results for $Auxiliary ContourNet_1$ and $Cascade ContourNet_2$ for visualization purpose.

\subsection{Qualitative Analysis}
\subsubsection{Detection with Contour}
\begin{figure}[!htb]
    \centering
 	\includegraphics[height=0.8in]{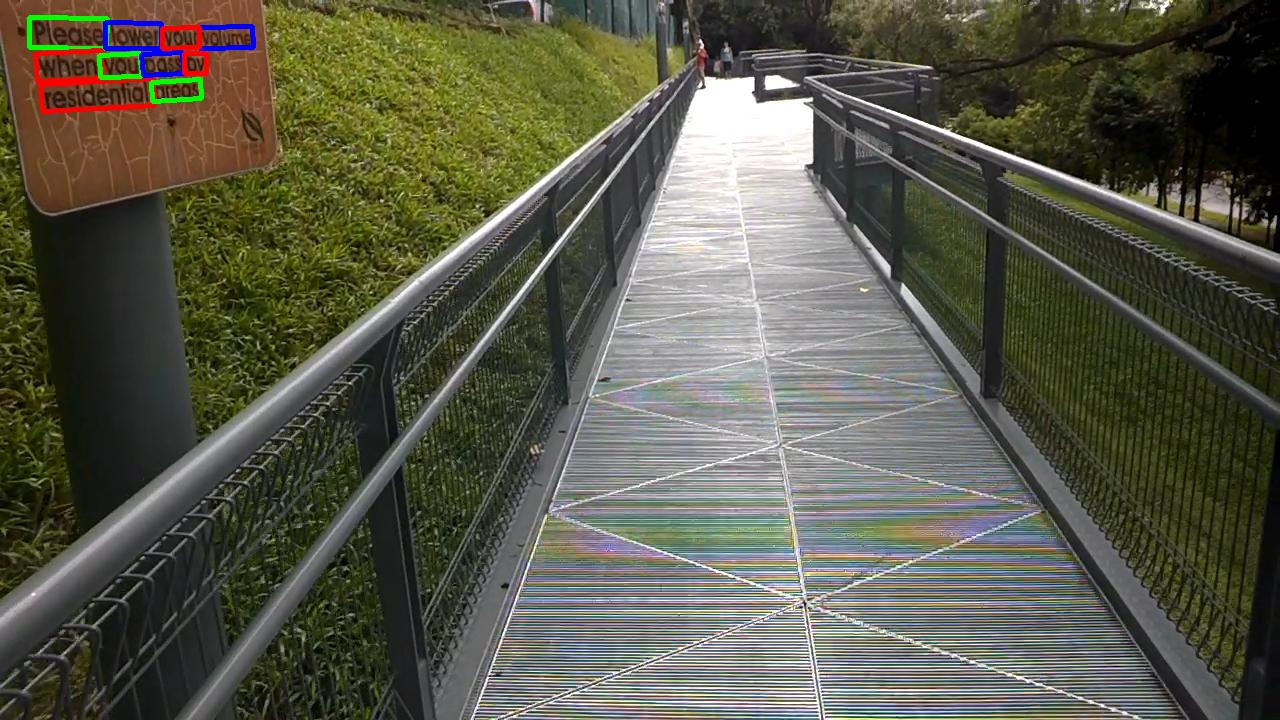}
	\includegraphics[height=0.8in]{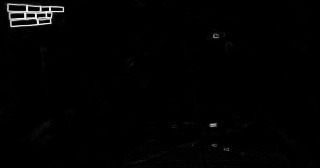}
	
	\includegraphics[height=0.8in]{Qualitative/img_170.jpg}
	\includegraphics[height=0.8in]{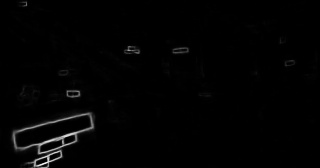}
	
	\includegraphics[height=0.74in]{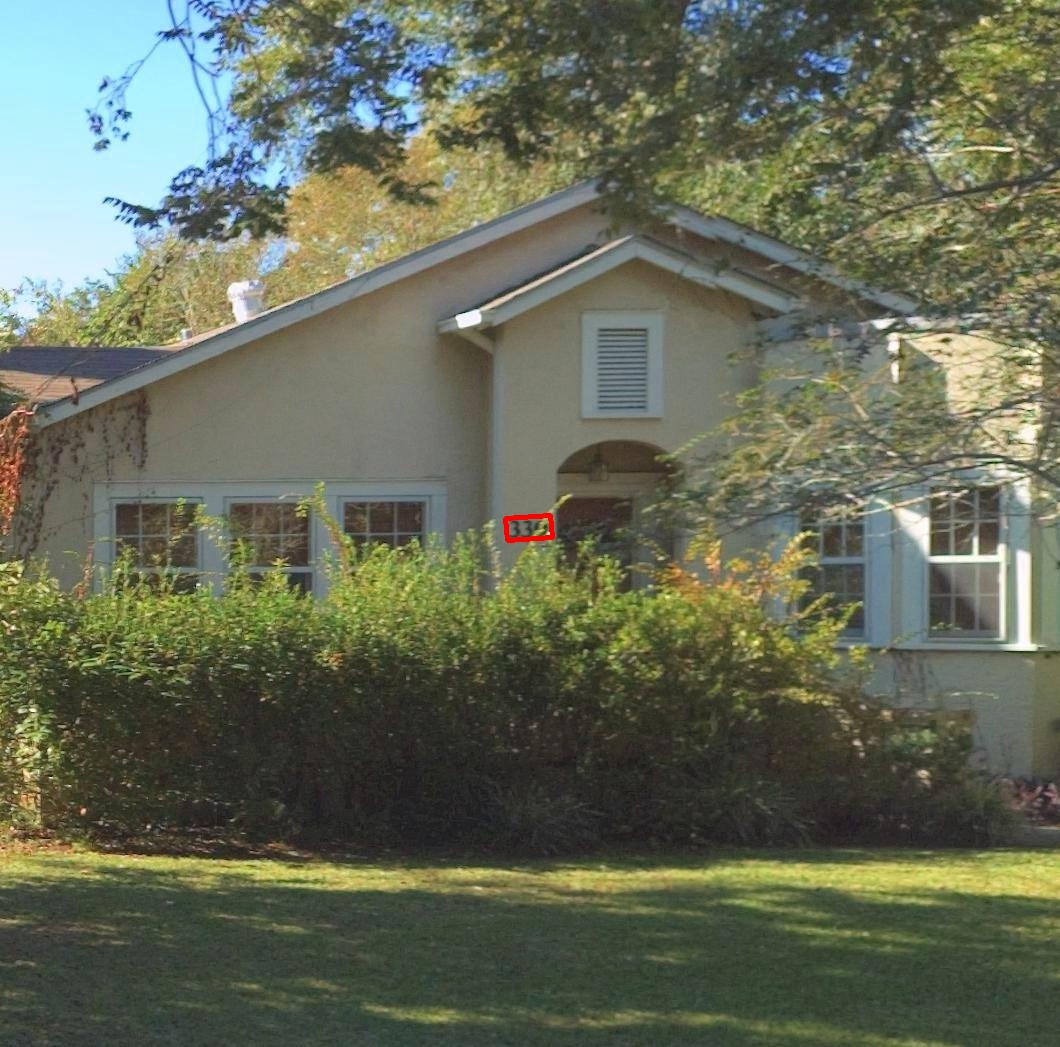}
	\includegraphics[height=0.74in]{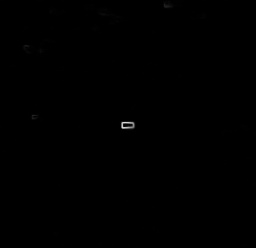}
	\includegraphics[height=0.74in]{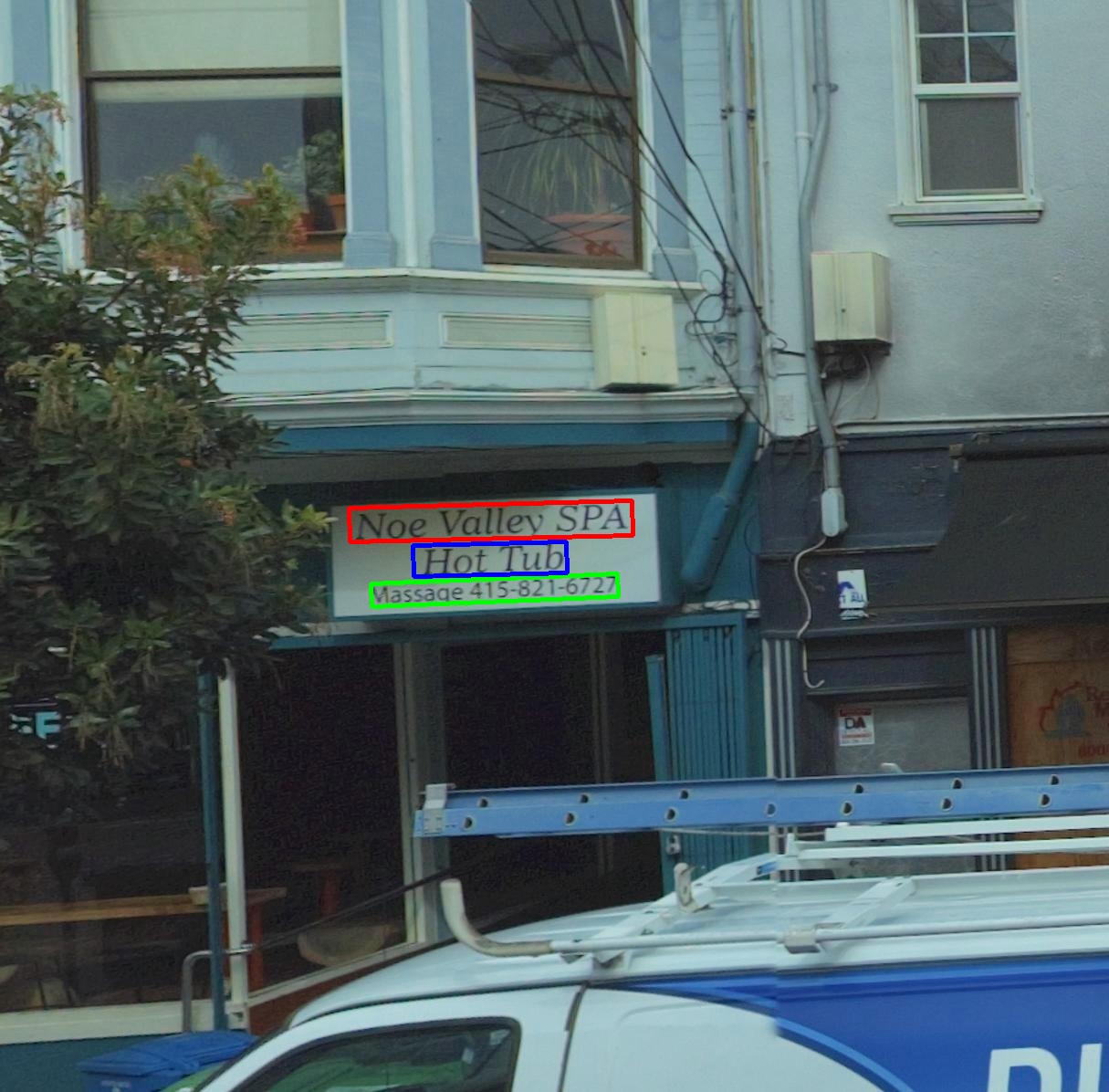}
	\includegraphics[height=0.74in]{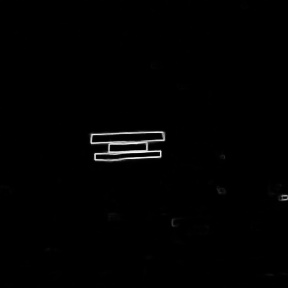}
 	\caption{Some detection results with their contour prediction. Better when zoom in.}
 	\label{fig:det_examples}
\end{figure}

Fig.~\ref{fig:det_examples} gives some example results with their contour predictions.
We can see that our model can effectively extract text instance for these cases.
Two things to be noted here:
(1) For uber text, sometimes the text instance is annotated as a line of text.
Our learned contour net can extract it effectively.
(2) Even when the text contour detection is not perfect(see the third image), by jointly learning the contour as well as the detection visual features, we can still give an accurate prediction.
This joint learning scheme differs from the idea~\cite{he2017single} as even though contour is not perfect, the detection results could still be good.
Note that all these visualized results are generated from regular training setting, not from the model in capacity study.

\subsubsection{Comparison with Baseline}
\begin{figure}[!htb]
    \centering
   	\includegraphics[height=0.58in]{comparison/ic15_img_14.jpg}
	\includegraphics[height=0.58in]{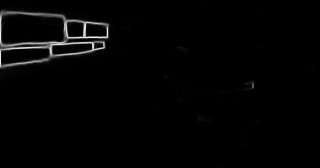}
	\includegraphics[height=0.58in]{comparison/ic15_img_14_baseline.jpg}
	
	\includegraphics[height=0.58in]{comparison/ic15_img_19.jpg}
	\includegraphics[height=0.58in]{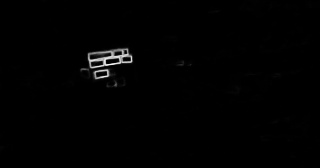}
	\includegraphics[height=0.58in]{comparison/ic15_img_19_baseline.jpg}
	
	\includegraphics[height=0.58in]{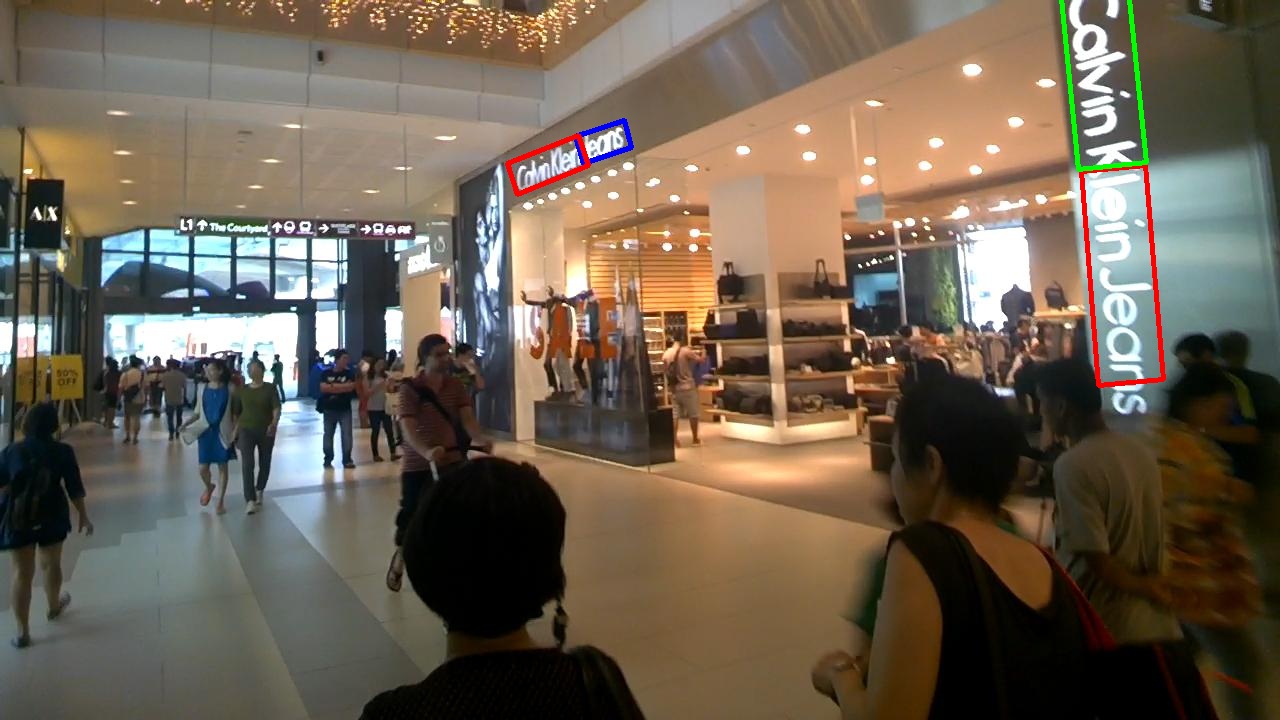}
	\includegraphics[height=0.58in]{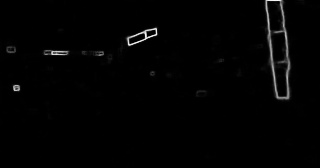}
	\includegraphics[height=0.58in]{comparison/ic15_img_42_baseline.jpg}
 	\caption{Comparison of the detection results from the proposed cascade contour model with baseline. For each column: (1) detection results from proposed model. (2) Predicted contour. (3) detection results from baseline model.}
 	\label{fig:comparison_egs}
\end{figure}

Fig.~\ref{fig:comparison_egs} shows the examples of detection results from contour cascade model compared with the baseline method.
We can see that the cascade model can give better instance predictions.

\subsubsection{Failure Examples}
\begin{figure}[!htb]
    \centering
   	\includegraphics[height=0.78in]{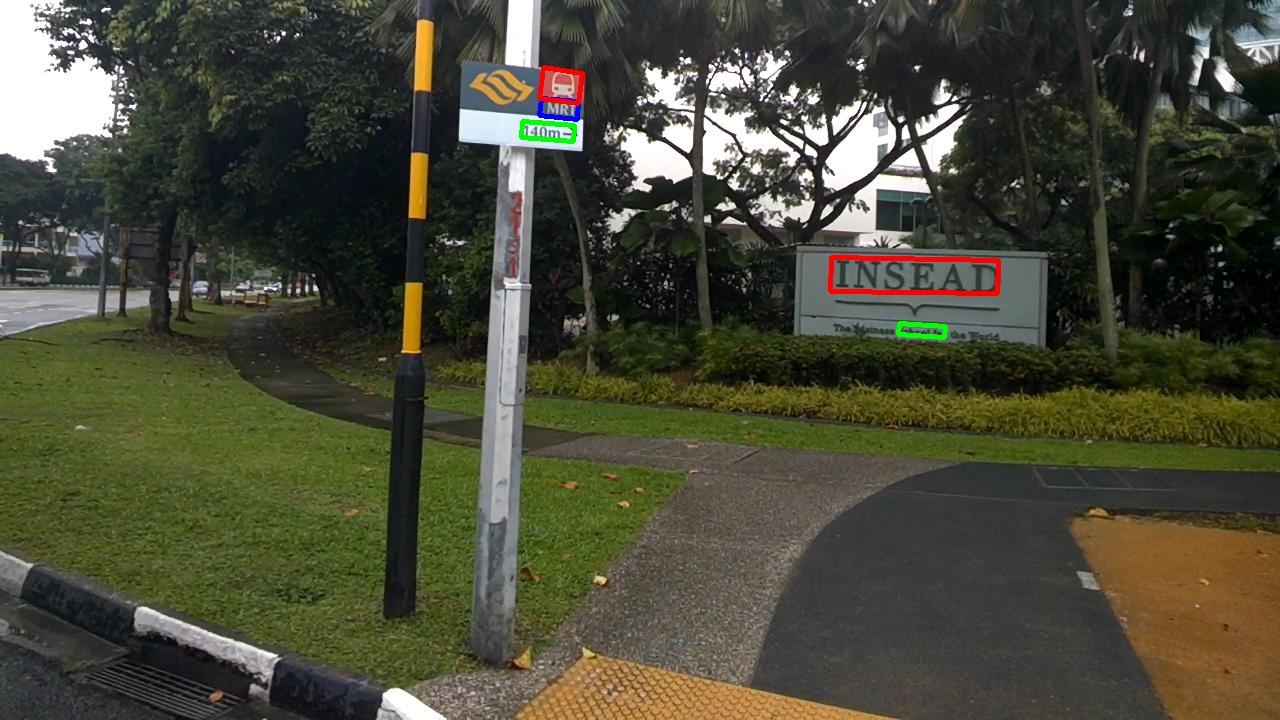}
	\includegraphics[height=0.78in]{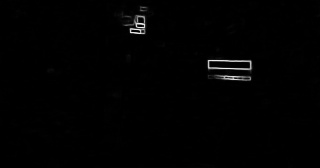}
	
   	\includegraphics[height=0.78in]{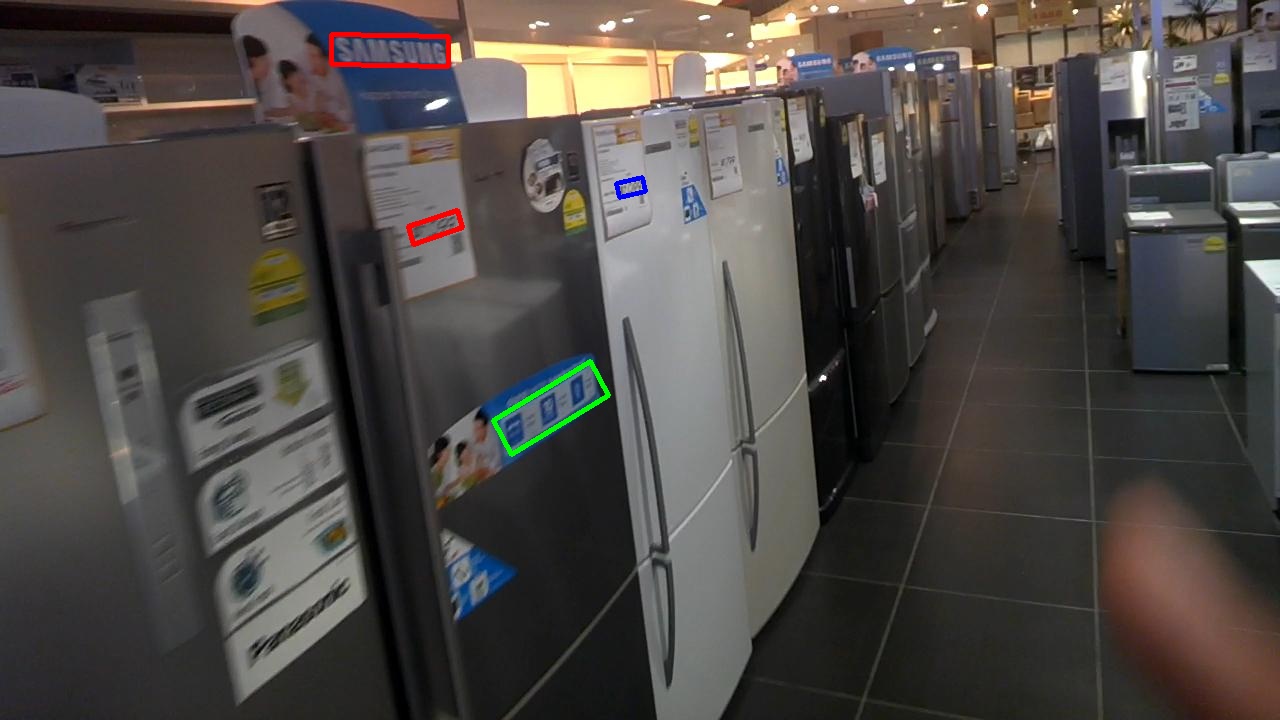}
	\includegraphics[height=0.78in]{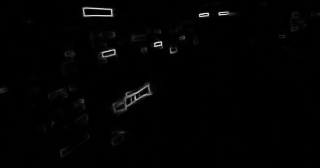}	
	
	\includegraphics[height=0.64in]{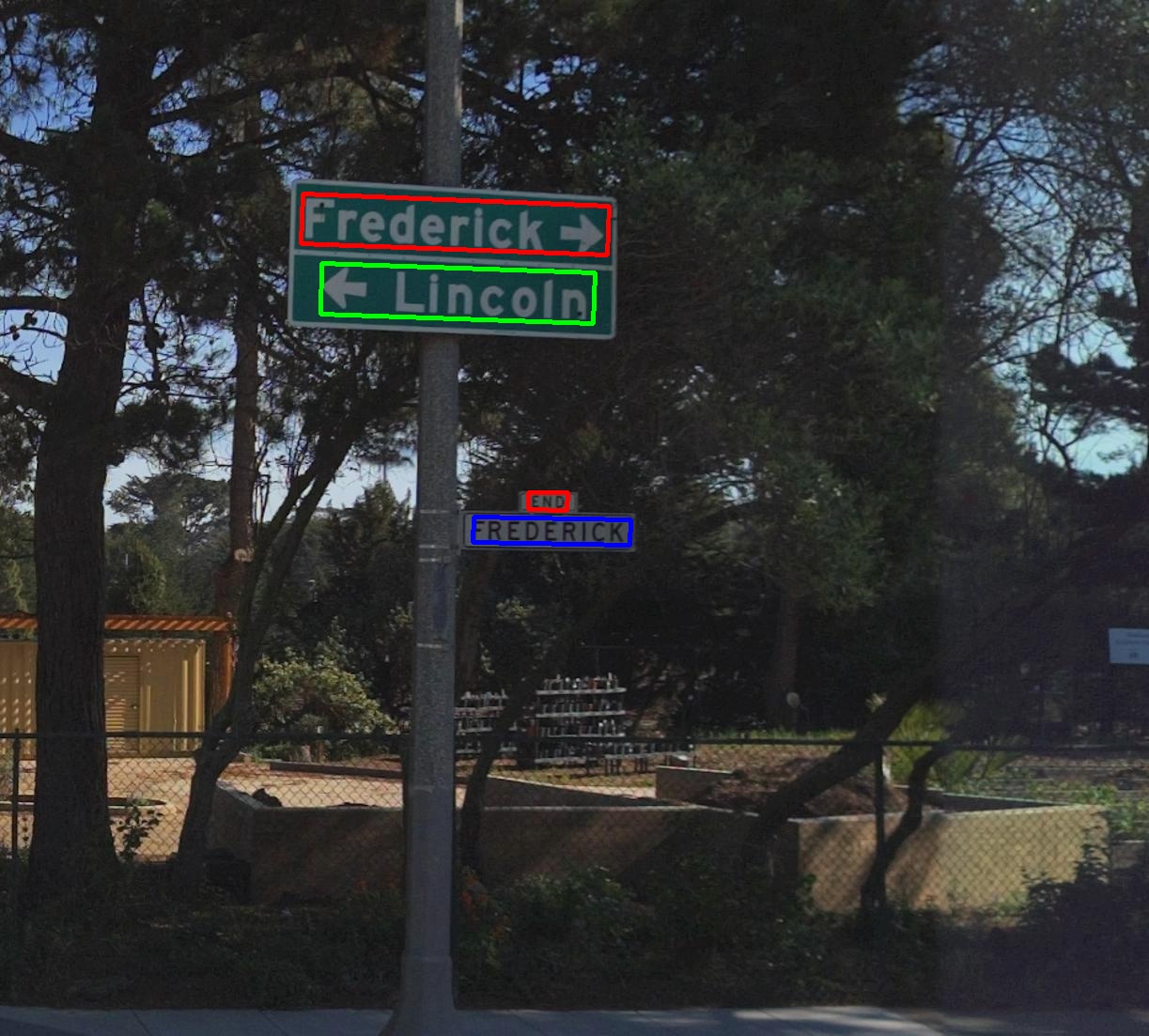}
	\includegraphics[height=0.64in]{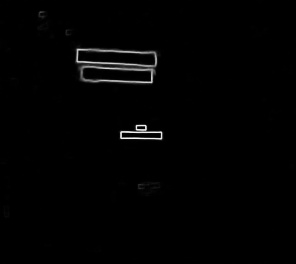}
	\includegraphics[height=0.64in]{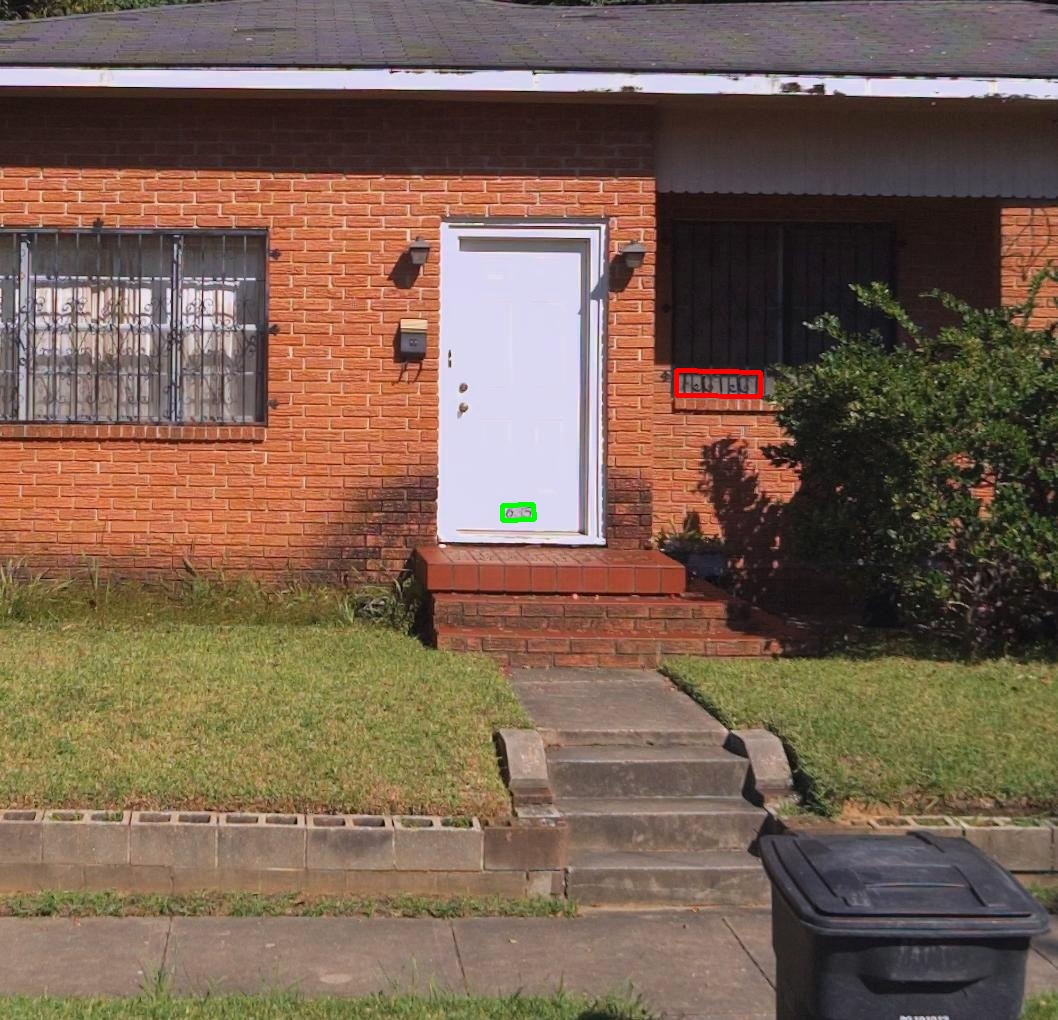}
	\includegraphics[height=0.64in]{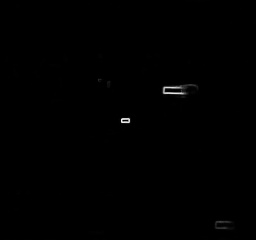}
 	\caption{Some imperfect or failure examples.}
 	\label{fig:failure_egs}
\end{figure}

Several imperfect or failure examples are in Fig.~\ref{fig:failure_egs}.
We also show the contour prediction.
We can see that some symbols are easily detected as text and will lead to imperfect predictions.
This is a common mistakes made by scene text detector.
When building an end-to-end pipeline with scene text reading, that could be removed partially by checking the transcribed text.

\section{Conclusion and Future Work}
We proposed a novel and effective framework for improving scene text detection.
The framework incorporates text instance contour segmentation to help improving scene text detection.
Unlike contour segmentation in regular images, we extend it specifically in the scene text area.
We now have a more semantic contour since it is not a subset of image edges, and it also contains instance level information.
Under such a definition, our framework doesn't need any extra annotation.
Traditional quadrilateral annotation is enough for training the model under our framework.
Model trained with the proposed framework has better capacity and generalizability.

\section{Acknowledgement}
This work was supported by NSF grant CCF 1317560 and a hardware grant from
NVIDIA.

{\small
\bibliographystyle{ieee}
\bibliography{egbib}
}

\end{document}